\documentclass[10pt,twocolumn,letterpaper]{article}

\usepackage[accsupp]{axessibility}

\usepackage{cvpr}              

\usepackage[dvipsnames]{xcolor}

\definecolor{cvprblue}{rgb}{0.21,0.49,0.74}
\usepackage[pagebackref,breaklinks,colorlinks,citecolor=cvprblue]{hyperref}

\usepackage{amsmath}
\usepackage{amsfonts}
\usepackage{amssymb}
\usepackage{wrapfig}
\usepackage{subcaption}
\usepackage{multirow}
 \usepackage{mathtools} 

\usepackage{verbatim}

\usepackage{anyfontsize}

\usepackage{microtype}
\usepackage{graphicx}
\usepackage{booktabs} 
\usepackage{multirow}
\usepackage{amsmath,amssymb}
\usepackage{booktabs}
\usepackage{caption,subcaption}

\usepackage{xcolor}
\definecolor{mygreen}{HTML}{3cb44b}
\definecolor{skyblue}{HTML}{beffff}
\definecolor{lightgreen}{HTML}{90ee90}

\usepackage{tcolorbox}
\usepackage{enumitem}
\setitemize{itemsep=10pt,topsep=0pt,parsep=0pt,partopsep=0pt}
\usepackage{adjustbox}

\newcommand{\RN}[1]{%
	\textup{\lowercase\expandafter{\it \romannumeral#1}}%
}
\usepackage{tabu}

\usepackage{amsmath}

\newcommand{\beq}{\vspace{0mm}\begin{equation}}
\newcommand{\eeq}{\vspace{0mm}\end{equation}}
\newcommand{\beqs}{\vspace{0mm}\begin{eqnarray}}
\newcommand{\eeqs}{\vspace{0mm}\end{eqnarray}}
\newcommand{\barr}{\begin{array}}
\newcommand{\earr}{\end{array}}

\usepackage{color, colortbl}
\definecolor{Gray}{gray}{0.93}

\usepackage{lipsum}

\usepackage{xcolor,amsmath}
\usepackage[linesnumbered,ruled,vlined]{algorithm2e}
\DontPrintSemicolon

\usepackage{color, colortbl}

\definecolor{emerald}{rgb}{0.31, 0.78, 0.37}
\definecolor{coralred}{rgb}{1.0, 0.25, 0.25}

\newcommand{\MyColorBox}[2][red]%
{%
    \settowidth{\Width}{#2}%
    \colorbox{#1}%
    {%
        \raisebox{-\DepthReference}%
        {%
                \parbox[b][\HeightReference+\DepthReference][c]{\Width}{\centering#2}%
        }%
    }%
}

\definecolor{codegray}{gray}{0.9}

\SetKwComment{Comment}{\color{green!50!black}\# }{}

\SetKwProg{Function}{def}{:}{}

\SetKwProg{For}{for}{:}{}
\SetKwProg{If}{if}{:}{}

\usepackage[T1]{fontenc}
\usepackage{helvet}

\usepackage{xcolor}
\usepackage{pifont}

\definecolor{lightblue}{rgb}{0.85, 0.95, 1}
\definecolor{lightorange}{rgb}{1, 0.95, 0.85}
\definecolor{lightpink}{rgb}{1, 0.9, 0.95}

\newcommand{\colorrect}[1]{\textcolor{#1}{\ding{110}}}

\newcommand{\shortname}{LLaVA}
\newcommand{\newshortname}{LLaVA-1.5}

\renewcommand{\paragraph}[1]{\vspace{1.25mm}\noindent\textbf{#1}}

\def\paperID{8} 
\def\confName{CVPR}
\def\confYear{2024}

\title{Improved Baselines with Visual Instruction Tuning}

\author{
Haotian Liu$^{1}$ \quad Chunyuan Li$^{2}$ \quad Yuheng Li$^{1}$ \quad Yong Jae Lee$^{1}$ \\
$^{1}$University of Wisconsin--Madison~~~~~~$^2$Microsoft Research, Redmond \\
\href{https://llava-vl.github.io}{https://llava-vl.github.io}
}

\begin{document}

\maketitle
\begin{abstract}
Large multimodal models (LMM) have recently shown encouraging progress with visual instruction tuning.
In this paper, we present the first systematic study to investigate the design choices of LMMs in a controlled setting under the LLaVA framework. We show that the fully-connected vision-language connector in LLaVA is surprisingly powerful and data-efficient.
With simple modifications to LLaVA, namely, using CLIP-ViT-L-336px with an MLP projection and adding academic-task-oriented VQA data with response formatting prompts, we establish stronger baselines that achieve state-of-the-art across 11 benchmarks.
Our final 13B checkpoint uses merely 1.2M publicly available data, and finishes full training in $\sim$1 day on a single 8-A100 node.
Furthermore, we present some early exploration of open problems in LMMs, including scaling to higher resolution inputs, compositional capabilities, and model hallucination, etc.
We hope this makes state-of-the-art LMM research more accessible. Code and model will be publicly available.
\end{abstract}
\section{Introduction}
\label{sec:intro}

Large multimodal models (LMMs) have become increasingly popular in the research community, as they are the key building blocks towards general-purpose assistants~\citep{li2023multimodal,gpt4v,alayrac2022flamingo}.
Recent studies on LMMs are converging on a central concept known as visual instruction tuning~\cite{llava}. The results are promising, \eg LLaVA~\cite{llava} and MiniGPT-4~\cite{zhu2023minigpt} demonstrate impressive results on natural instruction-following and visual reasoning capabilities. To better understand the capability of LMMs, multiple benchmarks~\cite{liu2023mmbench,yu2023mmvet,li2023seed,li2023pope,fu2023mme} have been proposed. Recent works further demonstrate improved performance by scaling up the pretraining data~\cite{dai2023instructblip,bai2023qwen,ye2023mplug}, instruction-following data~\cite{zhao2023svit,li2023mimic,dai2023instructblip,gong2023multimodal}, visual encoders~\cite{bai2023qwen}, or language models~\citep{lu2023empirical}, respectively. The LLaVA architecture is also leveraged in different downstream tasks and domains, including region-level~\cite{chen2023shikra,zhang2023gpt4roi} and pixel-level~\cite{lai2023lisa,wang2023visionllm} understanding, biomedical assistants~\cite{li2023llava}, image generation~\cite{black2023training}, adversarial studies~\cite{carlini2023aligned,zhao2023evaluating}.

\begin{figure}[t!]
\centering

\includegraphics[width=.9\linewidth]{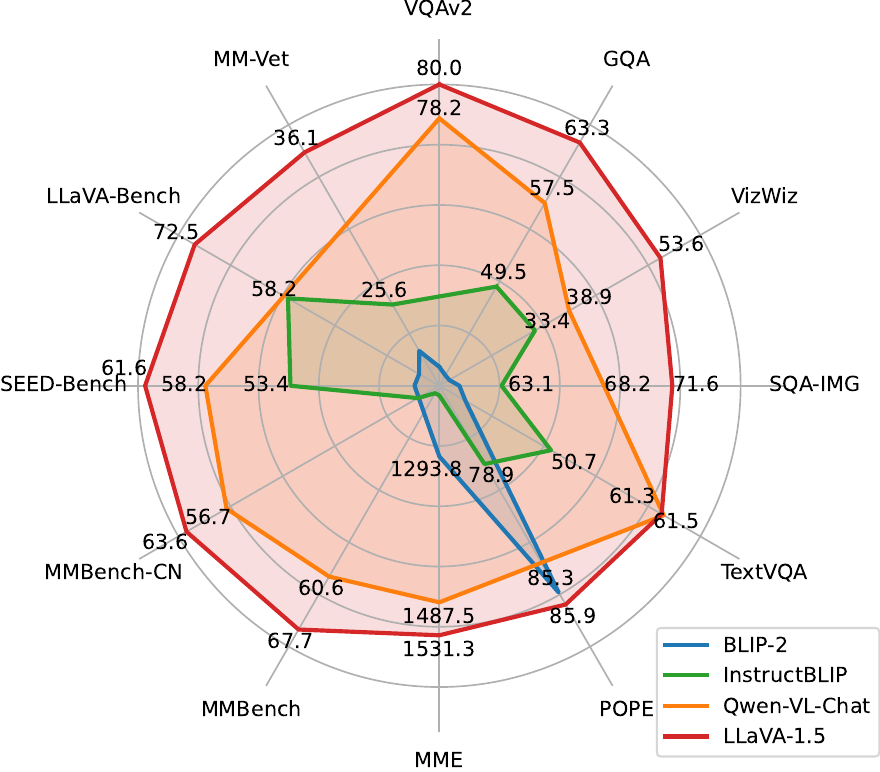} \\
\vspace{1mm}
\includegraphics[height=31mm]{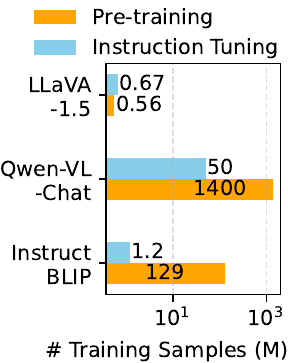}
\includegraphics[height=31mm]{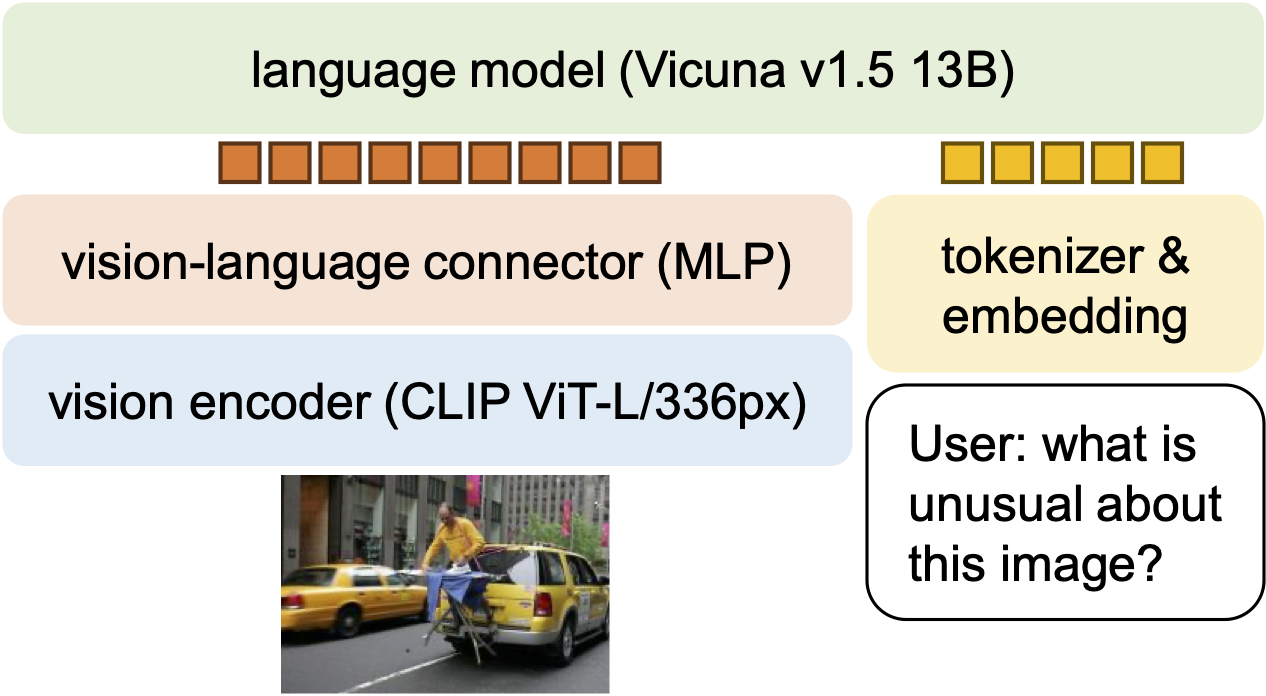} \\

\caption{\textbf{\newshortname{}} achieves SoTA on a broad range of 11 tasks (Top), with high training sample efficiency (Left) and simple modifications to \shortname{} (Right): an MLP connector and including academic-task-oriented data with response formatting prompts.}
\label{fig:architecture}
\end{figure}

However, despite many benchmarks and developments, it still remains unclear what the best recipe is to train LMMs towards the goal of general-purpose assistants. For example, LLaVA~\cite{llava} excels in conversational-style visual reasoning and even outperforms later approaches like InstructBLIP~\cite{dai2023instructblip} on such benchmarks~\cite{yu2023mmvet}, while InstructBLIP excels in traditional VQA benchmarks that demands single-word or short answers. Given the significant differences in the model architecture and training data between them, the root cause of the disparity in their capabilities remains elusive, despite conjectures~\cite{yu2023mmvet,liu2023mmbench}: the amount of training data, the usage of resamplers like Qformer~\cite{li2023blip}, \etc.
To this end, we present the first systematic study to investigate the design choices of LMMs in a controlled setting. Our study originates from  LLaVA and builds a road map by carefully making effective contributions from the perspectives of the input, model, and data.

First, we unveil that the fully-connected vision-language connector in LLaVA is surprisingly powerful and data-efficient, and we establish stronger and more feasible baselines built upon the LLaVA framework. We report that two simple improvements, namely, an MLP cross-modal connector and incorporating academic task related data such as VQA, are orthogonal to the framework of LLaVA, and when used with LLaVA, lead to better multimodal understanding capabilities. In contrast to InstructBLIP~\cite{dai2023instructblip} or Qwen-VL~\cite{bai2023qwen}, which trains specially designed visual resamplers on hundreds of millions or even billions of image-text paired data, LLaVA uses one of the simplest architecture design for LMMs and requires only training a simple fully-connected projection layer on merely 600K image-text pairs. Our final model can finish training in $\sim$1 day on a single 8-A100 machine and achieves state-of-the-art results on a wide range of benchmarks. Moreover, unlike Qwen-VL~\cite{bai2023qwen} that includes in-house data in training, \shortname{} utilizes only publicly available data.

Next, we delve into an early exploration of other open problems of large multimodal models. Our findings include: (1) Scaling to high-resolution image inputs. We show that LLaVA's architecture is versatile in scaling to higher resolutions by simply dividing images into grids and maintains its data efficiency; with the increased resolution, it improves the model's detailed perception capabilities and reduces hallucination. (2) Compositional capabilities. We find that large multimodal models are capable of generalizing to compositional capabilities. For example, training on long-form language reasoning together with shorter visual reasoning can improve the model's writing capability for multimodal questions. (3) Data efficiency. We show that randomly downsampling LLaVA's training data mixture by up to 75\% does not significantly decrease the model's performance, suggesting that the possibility of a more sophisticated dataset compression strategy can further improve LLaVA's already efficient training pipeline. (4) Data scaling. We provide empirical evidence for the scaling of data granularity in conjunction with the model's capability is crucial for an improved capability without introducing artifacts like hallucination.

In sum, we perform a systematic study on the training of large multimodal models, and introduce a simple yet effective approach to balance the multitask learning and effective scaling for large multimodal models. Our improved baselines, \newshortname{}, uses only \emph{public} data, achieves the state-of-the-art on a broad range of 11 tasks, and is significantly more data-efficient than previous approaches. By rethinking the conventional approaches and exploring the open problems in visual instruction tuning, we pave the way for more robust and capable systems for LMMs. We hope these improved and easily-reproducible baselines will provide a reference for future research in open-source LMMs.

\section{Related Work}
\label{sec:background}

\paragraph{Instruction-following large multimodal models (LMMs).}
Common architectures include a pre-trained visual backbone to encode visual features, a pre-trained large language model (LLM) to comprehend the user instructions and produce responses, and a vision-language cross-modal connector to align the vision encoder outputs to the language models. As shown in Fig.~\ref{fig:architecture}, LLaVA~\cite{llava} is perhaps the simplest architecture for LMMs. Optionally, visual resamplers (\eg Qformer~\cite{li2023blip}) are used to reduce the number of visual patches~\cite{zhu2023minigpt,dai2023instructblip,bai2023qwen}.
Training an instruction-following LMM usually follows a two-stage protocol.
First, the vision-language alignment pretraining stage leverages image-text pairs to align the visual features with the language model's word embedding space. Earlier works utilize relatively few image-text pairs (\eg $\sim$600K~\cite{llava} or $\sim$6M~\cite{zhu2023minigpt}), while some recent works pretrain the vision-language connector for a specific language model on a large amount of image-text pairs (\eg 129M~\cite{dai2023instructblip} and 1.4B~\cite{bai2023qwen}), to maximize the LMM's performance.
Second, the visual instruction tuning stage tunes the model on visual instructions~\cite{llava}, to enable the model to follow users' diverse requests on instructions that involve the visual contents. Dealing with higher resolution with grids in LMM are studied in con-current works~\cite{li2023otterhd,fuyu8b2024,ye2023ureader}.

\paragraph{Multimodal instruction-following data.}
In NLP, studies show that the quality of instruction-following data largely affects the capability of the resulting instruction-following models~\cite{zhou2023lima}. For visual instruction tuning, LLaVA~\cite{llava} is the pioneer to leverage text-only GPT-4 to expand the existing COCO~\cite{lin2014mscoco} bounding box and caption dataset to a multimodal instruction-following dataset that contains three types of instruction-following data: conversational-style QA, detailed description, and complex reasoning. LLaVA's pipeline has been employed to expand to textual understanding~\cite{zhang2023llavar}, million-scales~\cite{zhao2023svit}, and region-level conversations~\cite{chen2023shikra}. InstructBLIP~\cite{dai2023instructblip} incorporates academic-task-oriented VQA datasets to further enhance the model's visual capabilities. Conversely, \cite{chen2023visual} identifies that such naive data merging can result in models that tend to overfit to VQA datasets and thus are unable to participate in natural conversations. The authors further propose to leverage the LLaVA pipeline to convert VQA datasets to a conversational style. While this proves effective for training, it introduces added complexities in data scaling.
However, in NLP, the FLAN family~\cite{wei2021finetuned,chung2022scaling} shows that adding a large number of academic language tasks for instruction tuning can effectively improve the generalization ability. In light of this, we consider investigating the root cause of the inability to balance between natural conversations and academic tasks in multimodal models.

\section{Approach}
\label{sec:approach}



\subsection{Preliminaries}

As the seminal work of visual instruction tuning, LLaVA~\cite{llava} showcases commendable proficiency in visual reasoning capabilities, surpassing even more recent models on diverse benchmarks~\cite{yu2023mmvet,bitton2023visitbench} for real-life visual instruction-following tasks. LLaVA uses a single linear layer to project the visual features to language space, and optimizes the whole LLM for visual instruction tuning. However, LLaVA falls short on academic benchmarks that typically require short-form answers (\eg single-word), and tends to answer \emph{yes} for yes/no questions due to the lack of such data in the training distribution.

On the other hand, InstructBLIP~\cite{dai2023instructblip} is the pioneer to incorporate academic-task-oriented datasets like VQA-v2~\cite{goyal2017vqav2} along with LLaVA-Instruct~\cite{llava}, and demonstrates improved performance on VQA benchmarks. It pretrains Qformer~\cite{li2023blip} on 129M image-text pairs and only finetunes the instruction-aware Qformer for visual instruction tuning. However, recent studies~\cite{chen2023visual,yu2023mmvet} show that it does not perform as well as LLaVA on engaging in real-life visual conversation tasks. More specifically, as shown in Table \ref{subtab:instructblip_difficulty}, it can overfit to VQA training sets with short-answers, even on requests that require detailed responses.

\begin{table}[t!]
  \centering
  \begin{subtable}{\linewidth} 
    \centering 
    \scalebox{0.88}{
      \begin{tabular}{p{2.2cm} p{6.2cm} }
        \toprule
        \multicolumn{2}{l}{\bf Visual input example, Multitask Balancing Problem:}  \\
        \midrule
        &  \includegraphics[height=3cm]{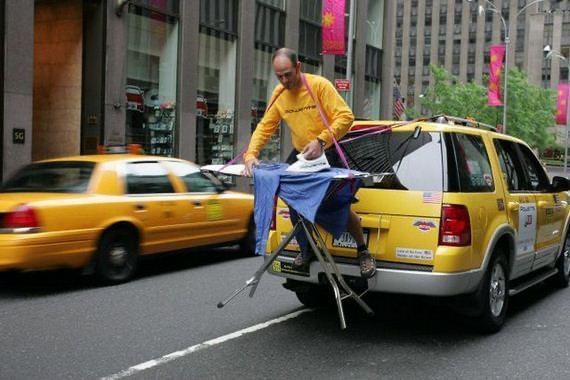} \\
        User & Is this unusual? Please explain in detail. \\
        InstructBLIP & yes \\
        \bottomrule
      \end{tabular}
    }
    \caption{Example of InstructBLIP~\cite{dai2023instructblip} (Vicuna-13B) having difficulty balancing between short- and long-form answers.}
    \label{subtab:instructblip_difficulty}  
  \end{subtable} 

  \vspace{3mm}
  
  \begin{subtable}{\linewidth} 
    \centering
    \scalebox{0.88}{
      \begin{tabular}{p{2.7cm} p{5.7cm} }
        \toprule
        \multicolumn{2}{l}{\bf Visual input example, Different Format Prompts:}  \\
        \midrule
        Normal prompt & What is the color of the shirt that the man is wearing? \\
        Response & The man is wearing a yellow shirt. \\
        \midrule
        Ambiguous prompt & Q: What is the color of the shirt that the man is wearing? A: \\
        Response & The man is wearing a yellow shirt. \\
        \midrule
        Formatting prompt & What is the color of the shirt that the man is wearing? \textbf{Answer the question using a single word or phrase.} \\
        Response & Yellow. \\
        \bottomrule
      \end{tabular}
    }
    \caption{Comparison of how different prompts regularize the output format. The results are obtained zero-shot directly after LLaVA undergoes the first-stage vision-language alignment pretraining, without the second-stage visual instruction tuning.}  
    \label{subtab:format_prompts_diff} 
  \end{subtable} 
  
  \caption{Visual input example to illustrate the challenge of (a) multitask balancing and (b) different format prompts. The same image input is used.}
  \label{tab:response_format_example}  
\end{table}

\subsection{Response Format Prompting}

We find that the inability~\cite{chen2023visual} to balance between short- and long-form VQA for approaches like InstructBLIP~\cite{dai2023instructblip}, which leverages instruction following data that includes both natural responses and short-answers, is mainly due to the following reasons.
First, \emph{ambiguous prompts on the response format}. For example, \emph{Q: \{Question\} A: \{Answer\}}. Such prompts do not clearly indicate the desired output format, and can overfit an LLM behaviorally to short-form answers even for natural visual conversations.
Second, \emph{not finetuning the LLM}. The first issue is worsened by InstructBLIP only finetuning the Qformer for instruction-tuning. It requires the Qformer's visual output tokens to control the length of the LLM's output to be either long-form or short-form, as in prefix tuning~\cite{li2021prefixtuning}, but Qformer may lack the capability of properly doing so, due to its limited capacity compared with LLMs like LLaMA.

Thus, to enable LLaVA to better handle short-form answers while addressing the issues of InstructBLIP, we propose to use a single response formatting prompt that clearly indicates the output format. It is appended at the end of VQA questions when promoting short answers: \emph{Answer the question using a single word or phrase}.
We find that when the LLM is \emph{finetuned} with such prompts, LLaVA is able to properly adjust the output format according to the user's instructions (see Table~\ref{subtab:format_prompts_diff}), and does not require additional processing of the VQA answers using ChatGPT~\cite{chen2023visual}, which further enables scaling to various data sources. As shown in Table~\ref{tab:scaling_ablation}, by merely including VQAv2~\cite{goyal2017vqav2} in training, LLaVA's performance on MME significantly improves (1323.8 \textit{vs} 809.6) and outperforms InstructBLIP by 111 points.

\begin{figure*}[t!]
\centering

\includegraphics[width=.8\linewidth]{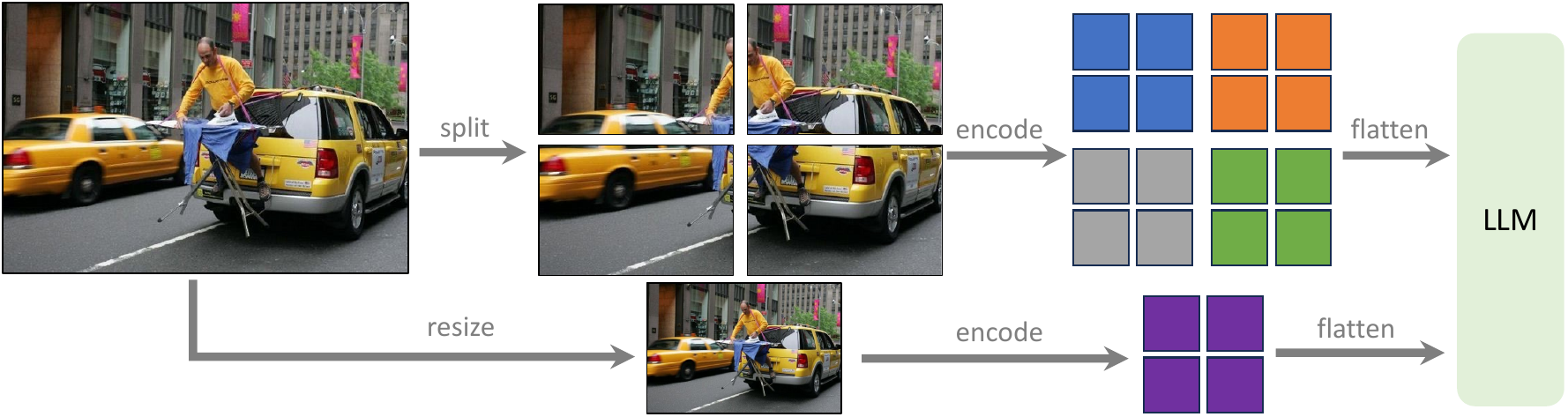} \\

\caption{\textbf{\newshortname{}-HD.} Scaling \newshortname{} to higher resolutions by splitting the image into grids and encoding them independently. This allows the model to scale to any resolution, without performing positional embedding interpolation for ViTs. We additionally concatenate the feature of a downsampled image to provide the LLM with a global context.}
\label{fig:high_res_arch}
\end{figure*}

\begin{table}[t!]
\centering
\scalebox{0.82}{
\begin{tabular}{l l p{6mm}p{7mm} | p{6mm}  p{8mm} c}
\toprule
\multicolumn{2}{l}{Method} & LLM & Res. & GQA & MME & MM-Vet \\
\midrule
\multicolumn{2}{l}{InstructBLIP} & 14B & 224 & 49.5 & 1212.8 & 25.6 \\
\midrule
\multicolumn{7}{l}{\it Only using a subset of InstructBLIP training data} \\
0 & \textbf{LLaVA} & 7B & 224 & -- & 809.6 & 25.5 \\
\rowcolor{lightblue} 1 & +VQA-v2 & 7B & 224 & 47.0 & 1197.0 & 27.7 \\
\rowcolor{lightblue} 2 &  +Format prompt & 7B & 224 & 46.8 & 1323.8 & 26.3 \\
\rowcolor{lightpink} 3 & +MLP VL connector & 7B & 224 & 47.3 & 1355.2 & 27.8 \\
\rowcolor{lightblue} 4 & +OKVQA/OCR & 7B & 224 & 50.0 & 1377.6 & 29.6 \\
\midrule
\multicolumn{7}{l}{\it Additional scaling} \\
\rowcolor{lightblue} 5 &  +Region-level VQA & 7B & 224 & 50.3 & 1426.5 & 30.8 \\
\rowcolor{lightorange} 6 &  +Scale up resolution & 7B & 336 & 51.4 & 1450 & 30.3 \\
\rowcolor{lightblue} 7 &  +GQA & 7B & 336 & 62.0$^*$ & 1469.2 & 30.7 \\
\rowcolor{lightblue} 8 &  +ShareGPT & 7B & 336 & 62.0$^*$ & 1510.7 & 31.1 \\
\rowcolor{lightpink} 9 &  +Scale up LLM & 13B & 336 & \textbf{63.3}$^*$ & \textbf{1531.3} & \textbf{36.1} \\
\bottomrule
\end{tabular}
}
\caption{
\textbf{Scaling results} on \colorrect{lightblue} data, \colorrect{lightpink} model, and \colorrect{lightorange} resolution. We choose to conduct experiments on GQA~\cite{hudson2019gqa}, MME~\cite{fu2023mme}, and MM-Vet~\cite{yu2023mmvet} to examine the representative capabilities of VQA with short answers, VQA with output formatting, and natural visual conversations, respectively.  $^*$Training images of GQA were observed during training.
}
\label{tab:scaling_ablation}
\end{table}

\subsection{Scaling the Data and Model}
\label{sec:scale_model_data}

\paragraph{MLP vision-language connector.}
Inspired by the improved performance in self-supervised learning by changing from a linear projection to an MLP~\cite{chen2020simple,mocov2}, we find that improving the vision-language connector's representation power with a two-layer MLP can improve LLaVA's multimodal capabilities, compared with the original linear projection.

\paragraph{Academic task oriented data.}
We further include additional academic-task-oriented VQA datasets for VQA, OCR, and region-level perception, to enhance the model's capabilities in various ways, as shown in Table~\ref{tab:scaling_ablation}. 
We first include four additional datasets that are used in InstructBLIP: open-knowledge VQA (OKVQA~\cite{okvqa}, A-OKVQA~\cite{schwenk2022okvqa}) and OCR (OCRVQA~\cite{mishra2019ocrvqa}, TextCaps~\cite{sidorov2020textcaps}). A-OKVQA is converted to multiple choice questions and a specific response formatting prompt is used: \emph{Answer with the option's letter from the given choices directly}. With only a subset of the datasets InstructBLIP uses, LLaVA already surpasses it on all three tasks in Table~\ref{tab:scaling_ablation}, suggesting LLaVA's effective design. Furthermore, we find further adding region-level VQA datasets (Visual Genome~\cite{krishna2017visual}, RefCOCO~\cite{kazemzadeh2014referitgame,mao2016generation}) improves the model's capability of localizing fine-grained visual details.

\paragraph{Additional scaling.}
We further scale up the input image resolution to 336$^2$ to allow the LLM to clearly ``see'' the details of images, by swapping the vision encoder to CLIP-ViT-L-336px (the highest resolution available for CLIP). In addition, we add the GQA dataset as an additional visual knowledge source. We also incorporate ShareGPT~\cite{sharegpt} data and scale up the LLM to 13B as in \cite{bai2023qwen,lu2023empirical,chen2023shikra}. Results on MM-Vet shows the most significant improvement when scaling the LLM to 13B, suggesting the importance of the base LLM's capability for visual conversations. 

\paragraph{\newshortname{}.}
We denote this final model with all the modifications as \newshortname{} (the last two rows in Table~\ref{tab:scaling_ablation}), which achieves an impressive performance that significantly outperforms the original LLaVA~\cite{llava}.

\paragraph{Computational cost.}
For \newshortname{}, we use the same pretraining dataset, and keep the training iterations and batch size roughly the same for instruction tuning as LLaVA~\cite{llava}. Due to the increased image input resolution to 336$^2$, the training of \newshortname{} is $\sim$2$\times$ as long as LLaVA: $\sim$6 hours of pretraining and $\sim$20 hours of visual instruction tuning, using 8$\times$ A100s.


\begin{table*}[t!]
\centering
\scalebox{0.8}{
\begin{tabular}{l l c c c | c c c c c }
\toprule
\multirow{2}{*}{Method} & \multirow{2}{*}{LLM} & Image & \multicolumn{2}{c|}{Sample Size} & VQAv2 & GQA & VisWiz & SciQA- & TextVQA \\
 & & Size & Pretrain & Finetune & \cite{goyal2017vqav2} & \cite{hudson2019gqa} & \cite{gurari2018vizwiz} & IMG~\cite{lu2022learn} & \cite{singh2019textvqa} \\
\midrule
BLIP-2~\cite{li2023blip} & Vicuna-13B & 224$^2$ & 129M & - & 65.0 & 41 & 19.6 & 61 & 42.5 \\
InstructBLIP~\cite{dai2023instructblip} & Vicuna-7B & 224$^2$ & 129M & 1.2M & -- & 49.2 & 34.5 & 60.5 & 50.1 \\
InstructBLIP~\cite{dai2023instructblip} & Vicuna-13B & 224$^2$ & 129M & 1.2M & -- & 49.5 & 33.4 & 63.1 & 50.7 \\
Shikra~\cite{chen2023shikra} & Vicuna-13B & 224$^2$ & 600K & 5.5M & 77.4$^*$ & -- & -- & -- & -- \\
IDEFICS-9B~\cite{idefics} & LLaMA-7B & 224$^2$ & 353M & 1M & 50.9 & 38.4 & 35.5 & -- & 25.9 \\
IDEFICS-80B~\cite{idefics} & LLaMA-65B & 224$^2$ & 353M & 1M & 60.0 & 45.2 & 36.0 & -- & 30.9 \\
Qwen-VL~\cite{bai2023qwen} & Qwen-7B & 448$^2$ & 1.4B$^\dagger$ & 50M$^\dagger$ & \underline{78.8}$^*$ & 59.3$^*$ & 35.2 & 67.1 & \textbf{63.8}$^*$ \\
Qwen-VL-Chat~\cite{bai2023qwen} & Qwen-7B & 448$^2$ & 1.4B$^*$ & 50M$^\dagger$ & 78.2$^*$ & 57.5$^*$ & 38.9 & 68.2 & \underline{61.5}$^*$ \\
\midrule
\rowcolor{Gray}
\textbf{\newshortname{}} & Vicuna-7B & 336$^2$ & \textbf{558K} & \textbf{665K} & \underline{78.5}$^*$ & \underline{62.0}$^*$ & \underline{50.0} & 66.8 & 58.2 \\
\rowcolor{Gray}
\textbf{\newshortname{}} & Vicuna-13B & 336$^2$ & \textbf{558K} & \textbf{665K} & \textbf{80.0}$^*$ & \textbf{63.3}$^*$ & \textbf{53.6} & \textbf{71.6} & \underline{61.3} \\
\rowcolor{Gray}
\textbf{\newshortname{}-HD} & Vicuna-13B & 448$^2$ & \textbf{558K} & \textbf{665K} & \textbf{81.8}$^*$ & \textbf{64.7}$^*$ & \textbf{57.5} & \underline{71.0} & \underline{62.5} \\
\midrule
\multicolumn{5}{l|}{\color{gray} Specialist SOTA: PaLI-X-55B~\cite{chen2023pali}} & \color{gray}86.1$^*$ & \color{gray}72.1$^*$ & \color{gray}70.9$^*$ & \color{gray}-- & \color{gray}71.4$^*$ \\
\bottomrule
\end{tabular}
}
\vspace{-1mm}
\caption{\textbf{Comparison with SoTA methods on academic-task-oriented datasets.} \newshortname{} achieves the best performance on 4/5 benchmarks, and ranks the second on the other. $^*$The training images/annotations of the datasets are observed during training. $^\dagger$Includes in-house data that is not publicly accessible.}
\label{tab:results}
\end{table*}

\begin{table*}[t!]
\centering
\scalebox{0.8}{
\begin{tabular}{l | c c c | c | c c | c c c | c | c }
\toprule
\multirow{2}{*}{Method} & \multicolumn{3}{c|}{POPE~\cite{li2023pope}} & MME & \multicolumn{2}{c|}{MMBench~\cite{liu2023mmbench}} & \multicolumn{3}{c|}{SEED-Bench~\cite{li2023seed}} & LLaVA- & MM-Vet \\
 & rand & pop & adv & \cite{fu2023mme} & en & cn & all & img & vid & Wild~\cite{llava} & \cite{yu2023mmvet}\\
\midrule
BLIP2-14B~\cite{li2023blip} & \textbf{89.6} & 85.5 & 80.9 & 1293.8 & -- & -- & 46.4 & 49.7 & 36.7 & 38.1 & 22.4 \\
InstructBLIP-8B~\cite{dai2023instructblip} & -- & -- & -- & -- & 36 & 23.7 & 53.4 & 58.8 & 38.1 & 60.9 & 26.2 \\
InstructBLIP-14B~\cite{dai2023instructblip} & \underline{87.7} & 77 & 72 & 1212.8 & -- & -- & -- & -- & -- & 58.2 & 25.6 \\
Shikra-13B~\cite{chen2023shikra} & -- & -- & -- & -- & 58.8 & -- & -- & -- & -- & -- & -- \\
IDEFICS-9B~\cite{idefics} & -- & -- & -- & -- & 48.2 & 25.2 & -- & 44.5 & -- & -- & -- \\
IDEFICS-80B~\cite{idefics} & -- & -- & -- & -- & 54.5 & 38.1 & -- & 53.2 & -- & -- & -- \\
Qwen-VL~\cite{bai2023qwen} & -- & -- & -- & -- & 38.2 & 7.4 & 56.3 & 62.3 & 39.1 & -- & -- \\
Qwen-VL-Chat~\cite{bai2023qwen} & -- & -- & -- & 1487.5 & 60.6 & 56.7 & \underline{58.2} & 65.4 & 37.8 & -- & -- \\
\midrule
\shortname{}-7B~\cite{llava} & 76.3 & 72.2 & 70.1 & 809.6 & 38.7 & 36.4 & 33.5 & 37.0 & 23.8 & 62.8 & 25.5 \\
\rowcolor{Gray}
\textbf{\newshortname{}-7B} & 87.3 & \underline{86.1} & \underline{84.2} & \underline{1510.7} & \underline{64.3} & 58.3 & \underline{58.6} & \underline{66.1} & 37.3 & 65.4 & \underline{31.1} \\
\rowcolor{Gray}
\textbf{\newshortname{}-13B} & 87.1 &  \textbf{86.2} & \textbf{84.5} & \textbf{1531.3} & \textbf{67.7} & \textbf{63.6} & \textbf{61.6} & \textbf{68.2} & \textbf{42.7} & \textbf{72.5} & \textbf{36.1} \\
\rowcolor{Gray}
\textbf{\newshortname{}-13B-HD} & 87.5 & \textbf{86.4} & \textbf{85.0} & 1500.1 & \textbf{68.8} & \underline{61.9} & \textbf{62.6} & \textbf{70.1} & \underline{41.3} & \underline{72.0} & \textbf{39.4} \\
\bottomrule
\end{tabular}
}
\vspace{-1mm}
\caption{\textbf{Comparison with SoTA methods on benchmarks for instruction-following LMMs.} \newshortname{} achieves the best overall performance.}
\label{tab:results}
\end{table*}

\subsection{Scaling to Higher Resolutions}

In Sec.~\ref{sec:scale_model_data}, we observe the advantage that scaling up the input image resolution improves the model's capabilities. However, the image resolution of the existing open source CLIP vision encoders is limited to 336$^2$, preventing the support of higher resolution images by simply replacing the vision encoder as we did in Sec.~\ref{sec:scale_model_data}. In this section, we present an early exploration of scaling the LMM to higher resolutions, while maintaining the data efficiency of \newshortname{}.

When using ViT~\cite{dosovitskiy2020image} as the vision encoder, to scale up the resolution, previous approaches mostly choose to perform positional embedding interpolation~\cite{bai2023qwen,li2023blip} and adapt the ViT backbone to the new resolution during finetuning. However, this usually requires the model to be finetuned on a large-scale image-text paired dataset~\cite{bai2023qwen,li2023blip}, and limits the resolution of the image to a fixed size that the LMM can accept during inference.

Instead, as shown in Fig.~\ref{fig:high_res_arch}, we overcome this by dividing the image into smaller image patches of the resolution that the vision encoder is originally trained for, and encode them independently. After obtaining the feature maps of individual patches, we then combine them into a single large feature map of the target resolution, and feed that into the LLM. To provide the LLM with the global context and to reduce the artifact of the split-encode-merge operation, we additionally concatenate the feature of a downsampled image to the merged feature map.
This allows us to scale the input to any arbitrary resolution and maintain the data efficiency of \newshortname{}. We call this resulting model \newshortname{}-HD.

\section{Empirical Evaluation}

\subsection{Benchmarks}

We evaluate \newshortname{} on a collection of both academic-task-oriented benchmarks and recent benchmarks specifically proposed for instruction-following LMMs, totalling 12 benchmarks.
For academic-task-oriented benchmarks, VQA-v2~\cite{goyal2017vqav2} and GQA~\cite{hudson2019gqa} evaluate model's visual perception capabilities on open-ended short answers.
VizWiz~\cite{gurari2018vizwiz} contains 8,000 images to evaluate model's zero-shot generalization on visual questions asked by visually impaired people. Following InstructBLIP~\cite{dai2023instructblip}, the image subset of ScienceQA~\cite{lu2022learn} with multiple choice are used to evaluate the zero-shot generalization on scientific question answering. TextVQA~\cite{singh2019textvqa} contains text-rich visual question answering.

For recent benchmarks proposed for instruction-following LMMs, POPE~\cite{li2023pope} evaluates model's degree of hallucination on three sampled subsets of COCO~\cite{lin2014mscoco}: random, common, and adversarial and we report the F1 score on all three splits. Other benchmarks evaluate the model's capabilities on a wide range of domains and applications, with different response formats. MME-Perception~\cite{fu2023mme} evaluates model's visual perception with yes/no questions. MMBench~\cite{liu2023mmbench} evaluates model's answer robustness with all-round shuffling on multiple choice answers. MMBench-CN~\cite{liu2023mmbench} is the Chinese-translated version of MMBench. SEED-Bench~\cite{li2023seed} evaluates model's performance on both images and videos with multiple choice, and we sample the frame in the middle to evaluate the accuracy on videos. LLaVA-Bench-in-the-Wild~\cite{llava} and MM-Vet~\cite{yu2023mmvet} evaluate model's capabilities in engaging in visual conversations on a diverse range of tasks, and evaluates the correctness and the helpfulness of the response with GPT-4 evaluation.

\begin{table}
  \begin{minipage}{0.99\linewidth}
\centering
\scalebox{0.80}{
\begin{tabular}{l p{7.5cm} }
\toprule
 \multicolumn{2}{l}{\bf Visual input example, Tricky Question:}  \\
\midrule
&  \includegraphics[height=3.5cm]{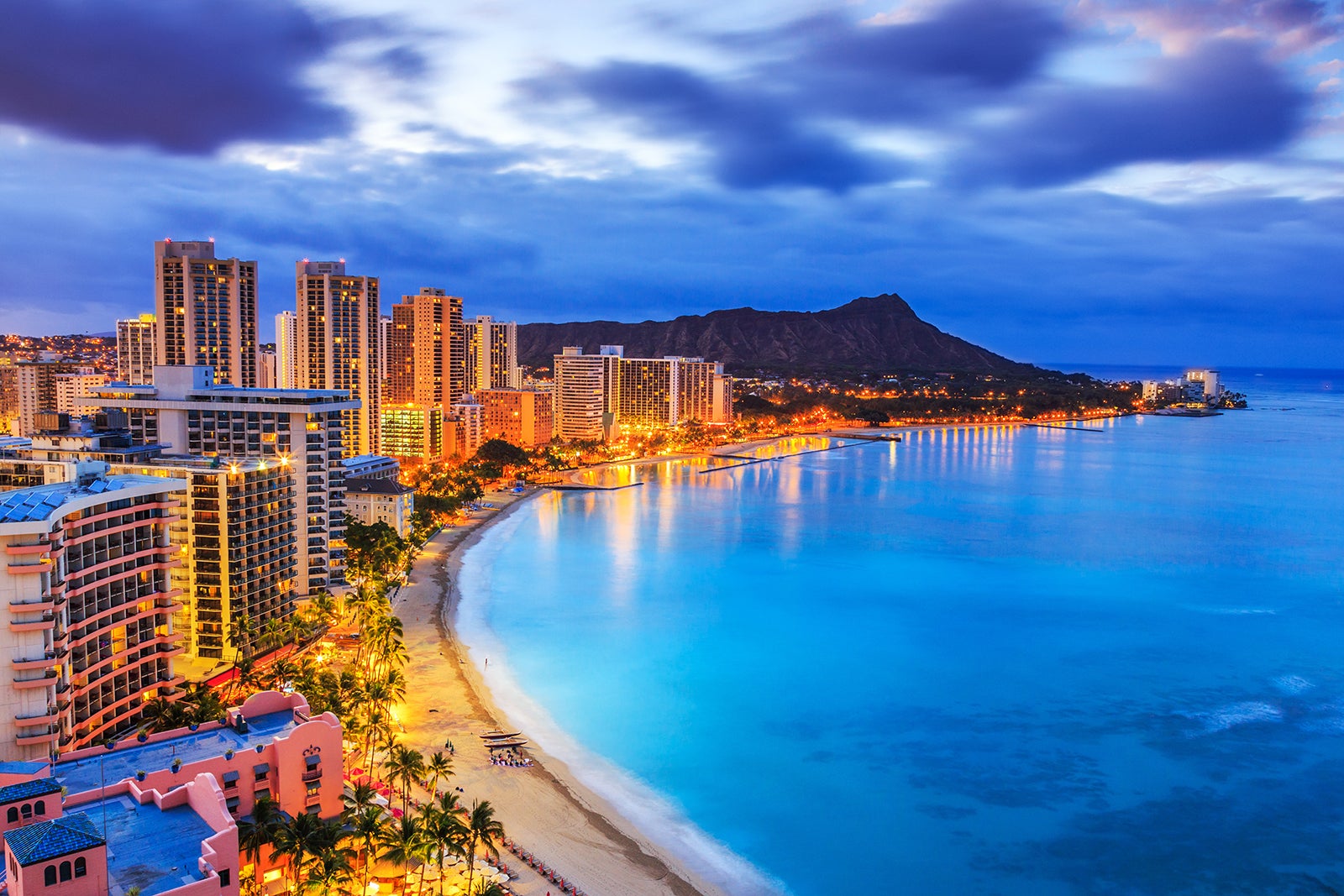} \\

User & If there are factual errors in the questions, point it out; if not, proceed to answering the question. \\
& What's happening in the desert? \\
\midrule
GPT-4V & Sorry, I cannot answer that question as there is no context about which desert you are referring to. \\ \midrule
\shortname{} & In the image, there is a city skyline with buildings and a beach with people enjoying the sunset. The scene is set in a desert, which is unusual for a beach setting. \\
\midrule
\newshortname{} & There are no deserts in the image. The image features a beach with palm trees, a city skyline, and a large body of water. \\
\bottomrule
\end{tabular}
}
\captionof{table}{\newshortname{} can detect and answer tricky questions when prompted to verify the question.}
\label{tab:tricky_example}  
  \end{minipage}
\end{table}

\subsection{Results}

We show that \newshortname{} achieves the best overall performance on 12 benchmarks, despite using magnitudes smaller pretraining and instruction tuning data compared with other methods~\cite{dai2023instructblip,bai2023qwen}. \newshortname{} significantly outperforms \shortname{} on all benchmarks for instruction-following LMMs. Note that it is challenging to evalute the original \shortname{} on academic datasets like VQA-v2~\cite{goyal2017vqav2} that demand open-ended short answers.

When we continue to scale up the image resolution to 448$^2$ with \newshortname{}-HD, it further improves the overall performance on all benchmarks, especially on tasks that require perception of details in the images (\eg OCR in MM-Vet, detailed description in LLaVA-Bench-in-the-Wild~\cite{llava}). Moreover, we find that adding the global context effectively recovers the model from the split-and-merge artifacts and guides the model to more easily locate the relevant regions from the high-resolution features (see appendix).

It is encouraging that \emph{\newshortname{} achieves the best performance with the simplest architecture, academic compute and public datasets, and yields a fully-reproducible and affordable baseline for future research}.
The results also suggest that visual instruction tuning plays an important role in improving an LMM's capabilities, and raises questions upon the common belief that LMMs require significant amount of vision-language alignment pretraining~\cite{dai2023instructblip,li2023blip,bai2023qwen}, despite that the vision encoders (\eg CLIP~\cite{radford2021learning}, OpenCLIP~\cite{openclip}, EVA-CLIP~\cite{fang2023eva}, \etc) are already pretrained on web-scale image-text paired data. \newshortname{} (even the 7B model) outperforms 80B IDEFICS~\citep{idefics}, a Flamingo-like LMM with billions of trainable parameters for cross-modal connection.
This also makes us rethink the benefits of the vision samplers and the necessity of the additional large-scale pretraining, in terms of multimodal instruction-following capabilities.

\begin{table}
  \begin{minipage}{0.99\linewidth}
\centering
\scalebox{0.8}{
\begin{tabular}{p{1.3cm} p{8.0cm} }
\toprule
 \multicolumn{2}{l}{\bf Visual input example, Constrained JSON Output:}  \\
\midrule
&  \includegraphics[height=3.5cm]{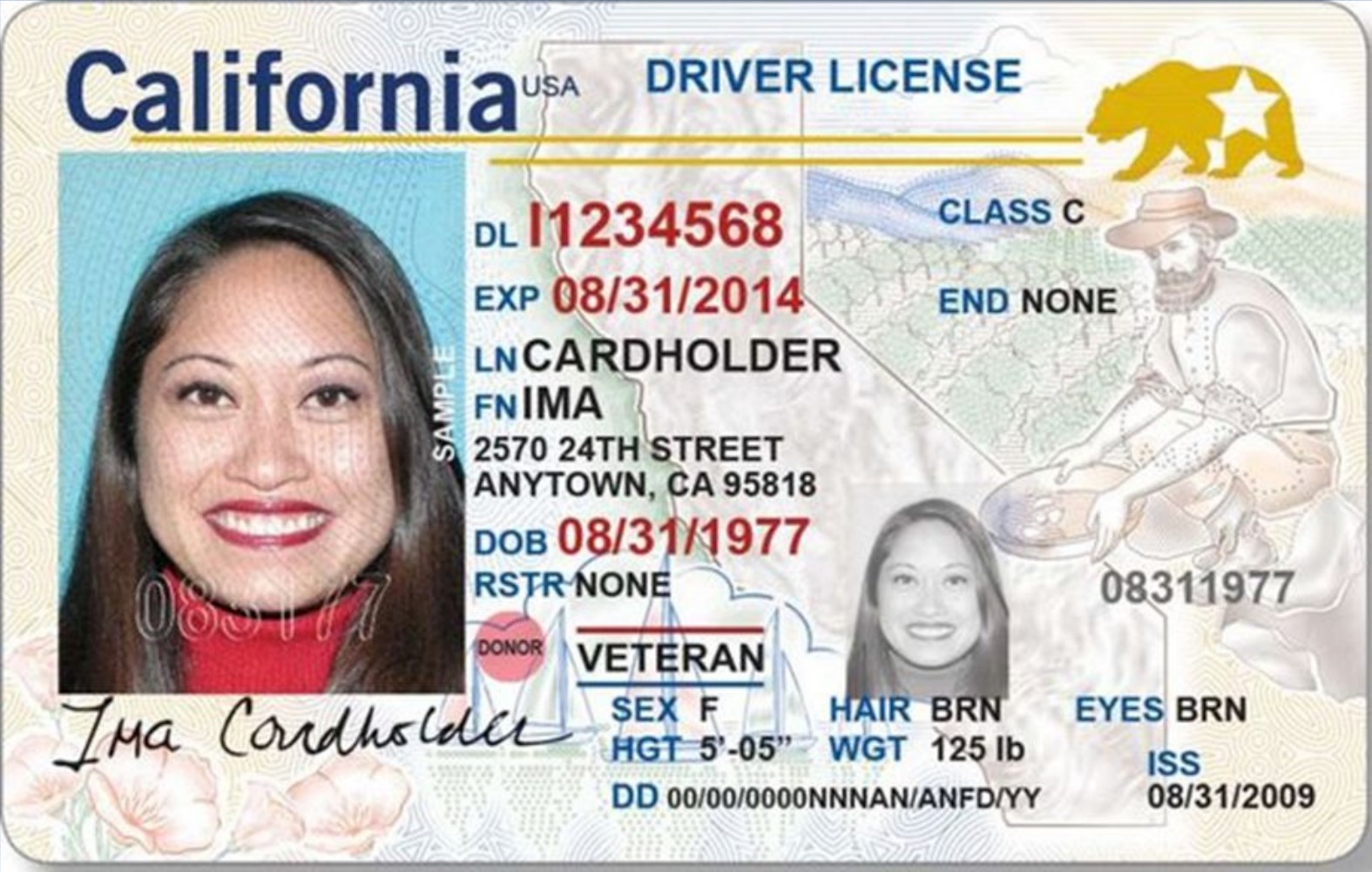} \\

User & Please read the text in this image and return the information in the following JSON format (note xxx is placeholder, if the information is not available in the image, put "N/A" instead). \\
& {\small \{"class": xxx, "DLN": xxx, "DOB": xxx, "Name": xxx, "Address": xxx, "EXP": xxx, "ISS": xxx, "SEX": xxx, "HGT": xxx, "WGT": xxx, "EYES": xxx, "HAIR": xxx, "DONOR": xxx\}} \\
\midrule
GPT-4V & {\small \{"class": "C", "DLN": "1234568", "DOB": "08/31/1977", "Name": "Ima Cardholder", "Address": "2570 24th Street, Anytown, CA 95818", "EXP": "08/31/2014", "ISS": "08/31/2011", "SEX": "F", "HGT": "5-05", "WGT": "125 lbs", "EYES": "BRN", "HAIR": "BRN", "DONOR": "VETERAN"\}} \\
\midrule
\shortname{} & {\small \{"class": "Driver License", "DLN": "CA123 456 789", "DOB": "01/01/1990", "Name": "Amy", "Address": "Anytown, CA", "EXP": "08/31/2023", "ISS": "Veteran", "SEX": “F", "HGT": "5'5", "WGT": "130", "EYES": "Brown", "HAIR": "Brown", "DONOR": "Unknown"\}} \\
\midrule
\newshortname{} & {\small \{"class": "C", "DLN": "1132456789", "DOB": "08/23/1971", "Name": "Ima Cardholder", "Address": "Anytown, State, Zip", "EXP": "08/23/2014", "ISS": "California", "SEX": "F", "HGT": "5'5", "WGT": "123", "EYES": "Brown", "HAIR": "Brown", "DONOR": "N/A"\}} \\
\bottomrule
\end{tabular}
}
\captionof{table}{\newshortname{} can extract information from the image and answer following the required format, despite a few errors compared with GPT-4V. GPT-4V results are obtained from \cite{yang2023dawn}.}
\label{tab:constrait_json}  
  \end{minipage}
\end{table}

\paragraph{Global context.}
For higher resolution, we pad and resize the image to a single image of 224$^2$, and concatenate it with the high resolution features to provide a global context. Ablation on a 7B model shows that the global context effectively boosts performance on all three validation benchmarks.

\begin{table}[h!]
\centering
\scalebox{0.82}{
\begin{tabular}{r | lll}
\toprule
 & GQA & MME & MM-Vet \\
\midrule
high-res patch only & 62.9 & 1425.8 & 31.9 \\
+global context & 63.8 {\bf \textcolor{emerald!80}{(+0.9)}} & 1497.5 {\bf \textcolor{emerald!80}{ (+71)}} & 35.1 {\bf \textcolor{emerald!80}{(+3.2)}} \\
\bottomrule
\end{tabular}
}
\label{tab:ablation_global_context}
\end{table}

\subsection{Emerging Properties}

\label{sec:emerging_property}

\paragraph{Format instruction generalization.}
Although \newshortname{} is only trained with a limited number of format instructions, it generalizes to others.
First, VizWiz~\cite{gurari2018vizwiz} requires the model to output ``Unanswerable'' when the provided content is insufficient to answer the question, and our response format prompt (see Appendix) effectively instructs the model to do so (11.1\% $\rightarrow$ 67.8\% on unanswerable questions).
We additionally present qualitative examples on instructing \newshortname{} to verify tricky questions (Fig.~\ref{tab:tricky_example}), respond in a constrained JSON format (Fig.~\ref{tab:constrait_json}), and more in appendix.

\paragraph{Multilingual multimodal capability.}
Though \newshortname{} is \emph{not} finetuned for multilingual multimodal instruction following \emph{at all} (all visual instructions including VQA are in English), we find that it is capable of following multilingual instructions.
This is partly due to the multilingual language instructions in ShareGPT~\cite{sharegpt}. Although ShareGPT does not contain images in its instructions, the model learns from this dataset the behavior of adaptively responding with the language that corresponds to the user's request. We empirically show that this behavior is transferred to visual conversations.
We also quantitatively evaluate the model's generalization capability to Chinese on MMBench-CN~\cite{liu2023mmbench}, where the questions of MMBench are converted to Chinese.
Notably, \newshortname{} outperforms Qwen-VL-Chat by +7.3\% (63.6\% vs 56.7\%), despite Qwen being finetuned on Chinese multimodal instructions while \newshortname{} is not.

\subsection{Ablation on LLM Choices}

\begin{figure}[t!]
\centering

\includegraphics[width=.99\linewidth]{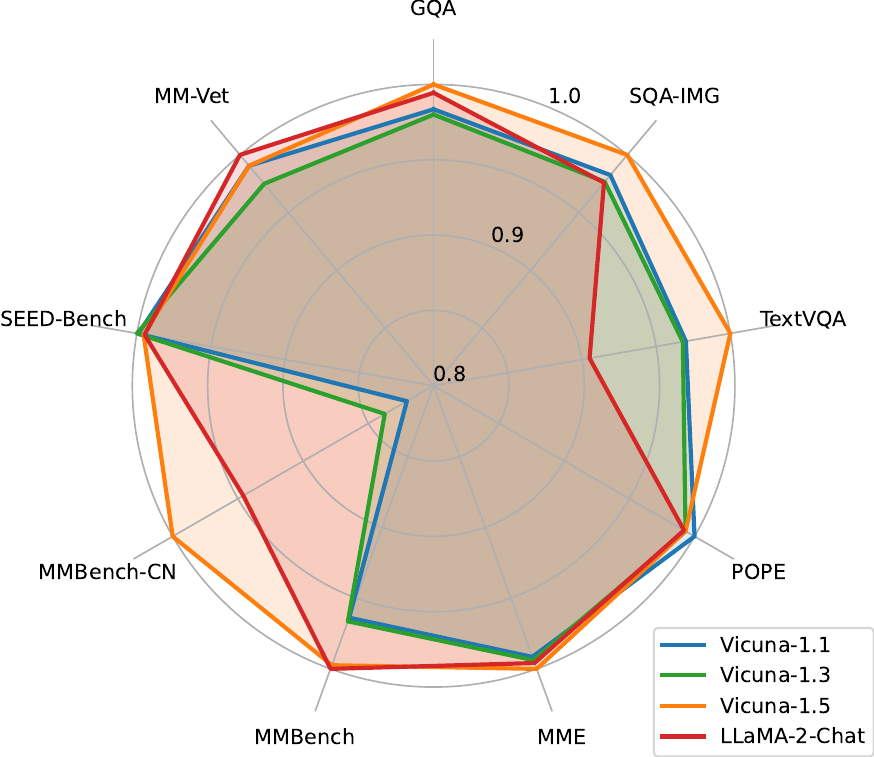} \\

\caption{\textbf{Ablation on LLM choices}. Data points represent the relative performance of the best performing variant for each dataset.}
\label{fig:ablate_llm_choice}
\end{figure}

In NLP, findings~\cite{touvron2023llama} suggest that the capability of the base LLM can affect its instruction-tuned successors. In this section, we explore two families of LLMs and study their contribution to the final model's multimodal capability: LLaMA-based (Vicuna-v1.1, Vicuna-v1.3) and LLaMA-2-based (Vicuna-v1.5, LLaMA-2-Chat). Vicuna-v1.3 and Vicuna-v1.5 use the same $\sim$150K ShareGPT~\cite{sharegpt} data (2$\times$ that used in v1.1). Unlike Vicuna series that is only trained with supervised instruction finetuning (SFT), LLaMA-2-Chat is further optimized with reinforcement-learning from human-feedback (RLHF). We visualize the relative performance of these variants in Fig.~\ref{fig:ablate_llm_choice}.

First, we find that Vicuna-v1.5 achieves the best overall performance, and LLaMA-2-based models generally perform better than LLaMA-1-based, suggesting the importance of the base language model. This is further evidenced by the results on MMBench-CN~\cite{liu2023mmbench}: despite Vicuna-v1.3 and v1.5 using the same ShareGPT data for instruction tuning, the performance in generalization to Chinese of Vicuna-v1.3 is significantly worse than v1.5.

Second, language instruction-tuning matters on specific capabilities that are required by each dataset. For example, although LLaMA-2-Chat and Vicuna-v1.5 achieves almost the same performance on MMBench, the generalization to MMBench-CN~\cite{liu2023mmbench} of LLaMA-2-Chat is worse than Vicuna-v1.5, which is partly due to that the most SFT/RLHF data of LLaMA-2-Chat is in English and does not contain as many multilingual data as in ShareGPT. Furthermore, TextVQA requires both the model's capability of identifying the text characters in the images, and also processing the noisy outputs from the OCR engine; such noise \emph{may} be more commonly observed in the ShareGPT data, which is collected in-the-wild from daily usage of ChatGPT.

\begin{figure}[t!]
\centering

\includegraphics[width=.99\linewidth]{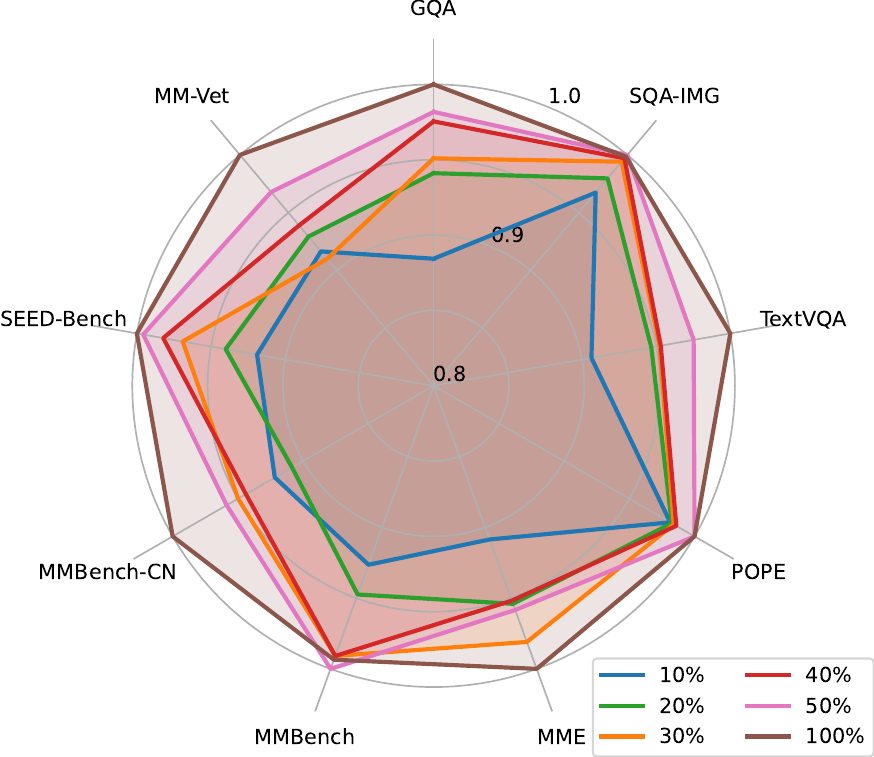} \\

\caption{\textbf{Ablation on data efficiency.}
Data points represent the relative performance of the best performing variant for each dataset.
}
\label{fig:ablate_data_efficiency}
\end{figure}

\section{Open Problems in LMMs}

Given the successful scaling of \newshortname{}, we conduct additional studies on open problems in LMMs using the model design and data mixture of \newshortname{}.

\subsection{Data Efficiency}

Despite the data efficiency of \newshortname{} when compared with approaches like InstructBLIP~\cite{dai2023instructblip}, the training of \newshortname{} still doubles when compared with \shortname{}. In this section, we conduct experiments for further improving the data efficiency by randomly sub-sampling the training data mixture of \newshortname{}, with a sampling ratio ranging from 0.1 to 0.5. We visualize the relative performance of different sampling variants in Fig.~\ref{fig:ablate_data_efficiency}.

First, the full data mixture provides the best knowledge coverage, and allows the model to achieve the best overall performance. To our surprise, with only 50\% of the samples, the model still maintains more than 98\% of the full dataset performance. This suggests that there is room for further improvements in data efficiency.

Second, when downsampling the dataset to 50\%, the model's performance on MMBench, ScienceQA, and POPE does not decrease at all, and it even slightly improves on MMBench. Similarly, the model's performance remains steady when further downscaling the data from 50\% to 30\%. These results show promise of having the less-is-more~\cite{zhou2023lima} benefit for multimodal models as well.

\subsection{Rethinking Hallucination in LMMs}

Hallucination is an important issue to tackle for LLMs and LMMs. Often in LMMs, we attribute the model's hallucination to the errors or hallucinations in the training dataset. For example, the detailed descriptions in LLaVA-Instruct~\cite{llava} may contain a small amount of hallucinated content, and it is believed that training on such data \emph{may} have caused the model to hallucinate when asked to ``describe the image in detail''.
However, we find that such hallucination is significantly reduced, when we scale the model's inputs to higher resolutions like 448$^2$.

This finding is interesting as it suggests that the LMMs may be robust to \emph{a few} such errors in the training data. However, when the input resolution is not sufficient for the model to discern all details in the training data, and the amount of data that is at that granularity beyond the model's capability becomes large enough, the model \emph{learns} to hallucinate.
This further suggests that there needs to be a balance between improving the data annotation with more details and the model's capability to properly process the information at such granularities.
We hope this finding provides a reference for future work in terms of dealing with hallucination and the scaling of the models and data.

\subsection{Compositional Capabilities}

We demonstrate interesting compositional capabilities in \newshortname{}: the model trained on a set of tasks independently generalizes to tasks that require a combination of these capabilities without explicit joint training. We note some of the findings below.

First, we observe an improved language capability in visual conversations after including the ShareGPT~\cite{sharegpt} data, including the multimodal multilingual capability as discussed in Sec.~\ref{sec:emerging_property}. Moreover, the model is more capable at providing longer and more detailed responses in visual conversations.
Second, the additional visual knowledge from the academic-task-oriented datasets, improves the visual groundness of \newshortname{}'s responses in visual conversations, as evidenced quantitatively by the improved results on MM-Vet~\cite{yu2023mmvet} and LLaVA-Wild~\cite{llava} in Table \ref{tab:results}.

However, there is still difficulty in achieving ideal performance for some tasks that require a certain combination of capabilities. For example, being able to correctly answer the attribute of a certain object in VQA, does not guarantee an accurate depiction of that object attribute in a detailed description of the whole image.
Furthermore, the capability of engaing in conversations with certain foreign languages (\eg Korean) still falls behind. See appendix for examples.

These findings suggest that the compositional capabilities of LMMs can be leveraged to improve the model's performance without significantly increasing the data by exhaustively including all task combinations. Yet, it can be further investigated, and a deeper understanding of the mechanism behind the compositional capabilities of LMMs can further improve the capability and the data efficiency of \newshortname{}.









\section{Conclusion}

In this paper, we take a step towards demystifying the design of large multimodal models, and propose a simple, effective, and data-efficient baseline, \newshortname{}, for large multimodal models. In addition, we explore the open problems in visual instruction tuning, scale LMMs to higher resolutions, and present some intriguing findings in terms of model hallucination and compositional capabilities for LMMs. We hope these improved and easily-reproducible baselines as well as the new findings will provide a reference for future research in open-source LMM.

\paragraph{Limitations.}
Despite the promising results demonstrated by \newshortname{}, it still has limitations including prolonged training for high-resolution images, lack of multiple-image understanding, limited problem solving capabilities in certain fields. It is not exempt from producing hallucinations, and should be used with caution in critical applications (\eg medical). See appendix for a detailed discussion.

\paragraph{Acknowledgements.}
This work was supported in part by NSF CAREER IIS2150012, and Institute of Information \& communications Technology Planning \& Evaluation(IITP) grants funded by the Korea government(MSIT) (No. 2022-0-00871, Development of AI Autonomy and Knowledge Enhancement for AI Agent Collaboration) and (No. RS-2022-00187238, Development of Large Korean Language Model Technology for Efficient Pre-training).



\section*{Appendix}
\appendix

This appendix is organized as follows.

\begin{itemize}
    \item In Section \ref{sec:appendix_implementation}, we show implementation details for \newshortname{}-HD (Sec.~\ref{sec:appendix_implementation_hd}), data and prompts (Sec.~\ref{sec:appendix_data}), and hyperparameters (Sec.~\ref{sec:appendix_hyperparameters}).
    \item In Section \ref{sec:appendix_qualitative}, we present more qualitative results for response format prompts (Sec.~\ref{sec:appendix_response_format_prompts}), compositional capabilities (Sec.~\ref{sec:appendix_compositional_capabilities}).
    \item In Section \ref{sec:limitation}, we discuss limitations with more details.
\end{itemize}

\section{Implementation Details}
\label{sec:appendix_implementation}




\subsection{\newshortname{}-HD}
\label{sec:appendix_implementation_hd}

\subsubsection{Preprocessing}

\paragraph{Overview.}
We use CLIP-ViT-L-14 (224$^2$) as the base image encoder. We first select and pad the input image to a target resolution that effectively captures its details, and split the image into 224$^2$ grids. All 224$^2$ image patches are encoded by the CLIP image encoder separately and their features are merged back to a single large feature map.  We then post-process the resulting feature map to a flattened list of features. We additionally concatenate the features of a fixed-resolution image to provide the model with a global context.

\paragraph{Target resolution selection.} We predefine a set of resolutions to support up to six grids (1x1, 1x2, 1x3, 1x4, 1x5, 1x6, 2x2, 2x3, and their transpose). This system allows for a maximum resolution of 672x448 (or 448x672). Two criteria are enforced in the target resolution selection: (1) \emph{Detail preservation}: the selected resolution preserves as much detail from the original image as possible; (2) \emph{Resource efficiency:} the resolution should not be excessively large to avoid unnecessary consumption of pixels and memory (\eg it should not select 448$^2$ for a 224$^2$ input image).

\paragraph{Postprocessing.}
We perform three steps of postprocessing to ensure that the final features can be processed effectively and efficiently by the language model.
(1) \emph{Padding removal.} Features corresponding exclusively to the paddings are discarded. This reduces the number of visual tokens processed by the language model and improves the efficiency.
(2) \emph{Row-end Tokens.} We append a special token to the end of each row of features, to provide an explicit indication of the shape of the image. Unlike the original \shortname{} and \newshortname{} that uses a fixed resolution, we now use a variable resolution for the image features of \newshortname{}-HD, such indication allows the language model to capture the exact shape and the size of the image for each sample.
(3) \emph{Flattening.} Finally, we flatten the image feature map and feed it into the language model along with language token features.



\subsubsection{Training}

Since we compute the visual features on the original 224$^2$ resolution that the vision encoder is trained on, we do not perform additional pretraining. We also do not perform additional high resolution pretraining for the visual projectors, and perform visual instruction tuning directly on the higher-resolution images.




\begin{table}[t!]
\centering
\scalebox{0.74}{
\begin{tabular}{p{20mm} l| p{70mm}}
\toprule
Data & Size & Response formatting prompts \\
\midrule
LLaVA~\cite{llava} & 158K & -- \\
ShareGPT~\cite{sharegpt} & 40K & -- \\
\midrule
VQAv2~\cite{goyal2017vqav2} & 83K & Answer the question using a single word or phrase. \\
GQA~\cite{hudson2019gqa} & 72K & \\
OKVQA~\cite{okvqa} & 9K & \\
OCRVQA~\cite{mishra2019ocrvqa} & 80K & \\
\midrule
A-OKVQA~\cite{schwenk2022okvqa} & 66K & Answer with the option's letter from the given choices directly. \\
\midrule
TextCaps~\cite{sidorov2020textcaps} & 22K & Provide a one-sentence caption for the provided image. \\
\midrule
RefCOCO & 48K & \emph{Note: randomly choose between the two formats} \\
\cite{kazemzadeh2014referitgame,mao2016generation} & & Provide a short description for this region. \\
\cmidrule{1-2}
VG~\cite{krishna2017visual} & 86K & Provide the bounding box coordinate of the region this sentence describes. \\
\midrule
Total & 665K & \\
\bottomrule
\end{tabular}
}
\caption{
\textbf{Instruction-following Data Mixture} of \newshortname{}.
}
\label{tab:data_mixture}
\end{table}

\begin{table}[t!]
\centering
\scalebox{0.74}{
\begin{tabular}{p{38mm}| p{65mm}}
\toprule
Data & Response formatting prompts \\
\midrule
LLaVA-Bench, MM-Vet & -- \\
\midrule
VQAv2, GQA, TextVQA, MME, POPE & Answer the question using a single word or phrase. \\
\midrule
ScienceQA, MMBench, SEED-Bench & Answer with the option's letter from the given choices directly. \\
\midrule
VizWiz & When the provided information is insufficient, respond with `Unanswerable'. Answer the question using a single word or phrase. \\
\bottomrule
\end{tabular}
}
\caption{
\textbf{Response format prompt} for evaluation.
}
\label{tab:data_mixture_eval}
\end{table}

\subsection{Data}
\label{sec:appendix_data}
Our final training data mixture contains a variety of datasets: VQA~\cite{goyal2017vqav2,hudson2019gqa,okvqa,schwenk2022okvqa}, OCR~\cite{mishra2019ocrvqa,sidorov2020textcaps}, region-level VQA~\cite{kazemzadeh2014referitgame,mao2016generation,krishna2017visual}, visual conversation~\cite{llava} and language conversation~\cite{sharegpt} data.
We adopt multiple strategies to reduce training cost and enhance efficiency, detailed as follows:
\begin{enumerate}
    \item For all VQA datasets, QA pairs from the same training image are merged into a single conversation.
    \item For ShareGPT~\cite{sharegpt}, we filter out invalid conversations as \cite{vicuna}. Unlike Vicuna~\cite{vicuna}, long conversations that surpass 2048 tokens are truncated rather than splitting to multiple conversations. This results in $\sim$40K conversations.
    \item Each QA pair in A-OKVQA~\cite{schwenk2022okvqa} is augmented $k$ times, where $k$ is the number of choices per question, to counterbalance the lack of multiple-choice data.
    \item 80K conversations are sampled from OCRVQA \cite{mishra2019ocrvqa}.
    \item For Visual Genome, we sample 10 annotations for images with additional annotations.
    \item For RefCOCO, conversations are dissected into segments, each containing fewer than 10 conversations.
    \item We obverse that language conversations are often longer than visual ones. For each batch, we sample conversations only from a single modality, and this speeds up the training by 25\%, and does not affect the final outcome.
\end{enumerate}

All data splits are concatenated together and sampled with the same  probability. We present the response formatting prompts of the final instruction-following data mixtures in Table~\ref{tab:data_mixture} and the response format prompts used for each evaluation benchmark in Table~\ref{tab:data_mixture_eval}.

\subsection{Hyperparameters}
\label{sec:appendix_hyperparameters}
The latest Vicuna v1.5~\cite{zheng2023judging} is used as the base LLM. \newshortname{} uses the same set of hyperparameters as the original \shortname{}, except that we halve the learning rate in pretraining due to the usage of the MLP projection layer instead of the original linear projection layer design.
We show the training hyperparameters for both first-stage vision-language alignment pretraining and the second-stage visual instruction tuning in Table~\ref{tab:hyperparameter}. We use greedy decoding for evaluation to ensure reproducibility.

\begin{table}[h!]
\centering
\scalebox{0.76}{
\begin{tabular}{l| c c}
\toprule
Hyperparameter & Pretrain & Finetune \\
\midrule
batch size & 256 & 128 \\
lr & 1e-3 & 2e-5 \\
lr schedule & \multicolumn{2}{c}{cosine decay} \\
lr warmup ratio & \multicolumn{2}{c}{0.03} \\
weight decay & \multicolumn{2}{c}{0} \\
epoch & \multicolumn{2}{c}{1} \\
optimizer & \multicolumn{2}{c}{AdamW} \\
DeepSpeed stage & 2 & 3 \\
\bottomrule
\end{tabular}
}
\caption{
\textbf{Hyperparameters} of \newshortname{} are the same as the original \shortname{}, except that we halve the learning rate in pretraining due to the usage of the MLP projection layer.
}
\label{tab:hyperparameter}
\end{table}

\section{Qualitative Results}
\label{sec:appendix_qualitative}

\subsection{Response Format Prompts}
\label{sec:appendix_response_format_prompts}

We show additional examples of \newshortname{} generalizing to different unseen response format prompts.

First, as shown in Table \ref{tab:format_prompt_generalization}, \newshortname{} can provide details at different granularities in response to user's requests. When requested by the user, it is also capable of switching between response formats within the conversations.

Second, we provide another example of the constrained prompting to generate the prompts for Stable Diffusion models. We show an example of generating anime prompts in Table \ref{tab:anime_prompt_generation}.

\begin{table}[t!]
  \begin{minipage}{0.99\linewidth}
\centering
\scalebox{0.87}{
\begin{tabular}{p{1.7cm} p{6.7cm} }
\toprule
 \multicolumn{2}{l}{\bf Visual input example, Different Format Prompts:}  \\
\midrule
&  \includegraphics[height=3cm]{figs/extreme_ironing.jpg} \\

User & What is the color of the shirt that the man is wearing? \textbf{Answer the question using a single word or phrase.} \\
\newshortname{} & Yellow \\
\midrule
User & Is this unusual? \textbf{Please explain briefly with a short sentence.} \\
\newshortname{} & Yes, it is unusual to see a man ironing clothes while standing on the back of a moving car. \\
\midrule
User & Is this unusual? \textbf{Please explain in detail.} \\
\newshortname{} & Yes, this is an unusual scene. A man is ironing clothes while standing on the back of a moving yellow SUV. This is not a typical activity one would expect to see on a city street, as ironing clothes is usually done indoors, in a more private setting. The man's actions are not only unconventional but also potentially dangerous, as standing on the back of a moving vehicle can be hazardous and may lead to accidents or injuries. Additionally, the presence of a suitcase and a handbag in the scene suggests that the man might be traveling or on the go, which further emphasizes the unusual nature of the situation. \\
\midrule
User & What is unusual about this image? \textbf{Answer briefly with a single sentence.} \\
\newshortname{} & A man is ironing clothes in the back of a car. \\
User & \textbf{Give a more detailed explanation.} \\
\newshortname{} & The image shows a man ironing clothes in the back of a car, which is an unusual sight as ironing is typically done indoors. The man is standing on a folding table, and the car appears to be a taxi. The scene takes place on a city street, with other cars and a truck visible in the background. \\
\bottomrule
\end{tabular}
}
\captionof{table}{\newshortname{} learns to format the response according to the user's request, generalizes to unseen format instructions, and can alter the response format within the conversation upon the user's request.}
\label{tab:format_prompt_generalization}
  \end{minipage}
\end{table}

\clearpage

\begin{table*}[t!]
  \begin{minipage}{0.99\linewidth}
\centering
\scalebox{0.88}{
\begin{tabular}{p{1.7cm} p{17.1cm} }
\toprule
 \multicolumn{2}{l}{\bf Visual input example, Visual Writing Task:}  \\
\midrule
&  \includegraphics[height=3cm]{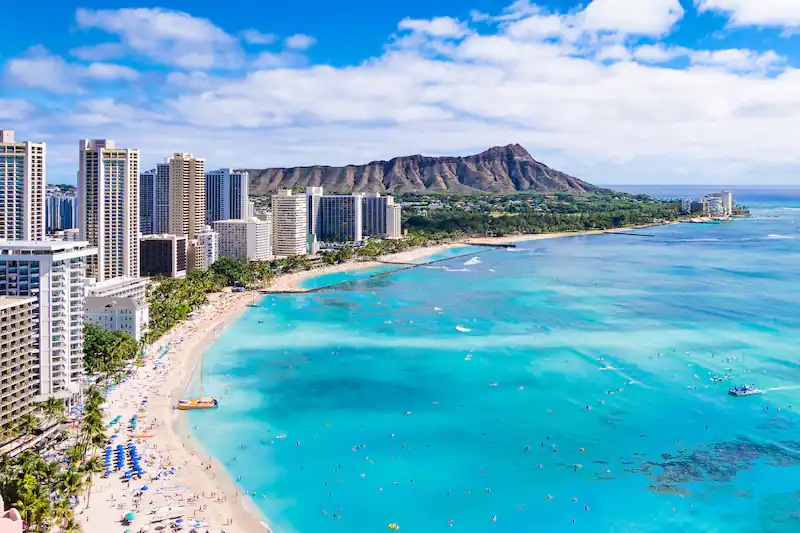} \\

User & Compose an engaging travel blog post about a recent trip to \textbf{this place}, highlighting \textbf{cultural experiences} and \textbf{must-see attractions}. \\
\midrule
\shortname{} & In my recent trip to Honolulu, I had the opportunity to experience the vibrant local culture and explore the beautiful beaches that this city has to offer. One of the must-see attractions was Waikiki Beach, where I witnessed a lively atmosphere with people sunbathing, swimming, and playing beach games. The crystal blue waters of the Pacific Ocean were a stunning sight, and the white sandy beach provided the perfect spot for relaxation. \\
 & Aside from the beach, I also had the chance to immerse myself in the local culture by visiting the Pearl Harbor Memorial, which paid tribute to the events that led to the United States' involvement in World War II. The memorial was a somber reminder of the past, and it was fascinating to learn about the history surrounding the attack on Pearl Harbor. \\
 & Other cultural experiences included trying local Hawaiian cuisine, such as poke bowls and traditional Hawaiian dishes, which were delicious and offered a unique taste of the region. I also had the opportunity to explore the city's bustling streets, where I came across local shops, markets, and street performers, adding to the lively atmosphere. \\
 & Overall, my trip to Honolulu was an enriching experience, filled with cultural discoveries, breathtaking natural beauty, and unforgettable memories. \\
\midrule
\newshortname{} & Title: A Sunny Escape to the Tropical Paradise of Waikiki Beach \\
\\
 & Introduction: \\
 & I recently had the pleasure of visiting the beautiful Waikiki Beach in Hawaii, and I must say, it was an unforgettable experience. The crystal blue waters, the lively atmosphere, and the stunning cityscape in the background made it a perfect destination for a relaxing getaway. In this travel blog post, I'll share my experiences and must-see attractions during my trip to Waikiki Beach. \\
 \\
 & Day 1: Arrival and Exploration \\
 & Upon arrival, I was greeted by the warm and welcoming Hawaiian culture. The vibrant colors and lively music set the tone for my adventure. I decided to explore the area around Waikiki Beach, starting with the iconic Diamond Head Crater. The hike to the top was a bit challenging, but the breathtaking views of the beach and the city made it worthwhile. \\
 \\
 & Day 2: Beach Day and Cultural Experiences \\
 & I woke up early and headed to the beach to enjoy a day of sunbathing and swimming. The beach was bustling with activity, and I loved watching the surfers catching waves. In the evening, I attended a traditional Hawaiian luau, where I learned about the local customs and traditions, including hula dancing and the art of making leis. \\
 \\
 & Day 3: City Tour and Sunset Cruise \\
 & On the third day, I took a guided city tour to learn more about the history and culture of Hawaii. I visited the Pearl Harbor Memorial, the Bishop Museum, and the Iolani Palace, which was once the residence of the Hawaiian monarchs. In the evening, I went on a sunset cruise around the Waikiki coastline, where I enjoyed a delicious meal and witnessed the stunning sunset over the ocean. \\
 \\
 & Day 4: Snorkeling and Relaxation \\
 & For my last day in Waikiki, I decided to go snorkeling in the crystal clear waters of Hanauma Bay. The marine life was incredible, and I had the opportunity to see colorful fish and coral reefs up close. After a long day of exploring, I relaxed on the beach and soaked up the sun. \\
 \\
 & Conclusion: \\
 & My trip to Waikiki Beach was an unforgettable experience filled with adventure, culture, and relaxation. From the breathtaking views of the ocean to the vibrant cityscape, this tropical paradise has something for everyone. I highly recommend visiting Waikiki Beach for a sunny escape and a chance to immerse yourself in the beautiful Hawaiian culture.
\\
\bottomrule
\end{tabular}
}
\captionof{table}{\newshortname{} provides more detailed, visually-grounded responses for writing tasks with visual inputs than \shortname{}.}
\label{tab:visual_writing_task}
  \end{minipage}
\end{table*}

\begin{figure*}[t!]
\centering

\includegraphics[width=.99\linewidth]{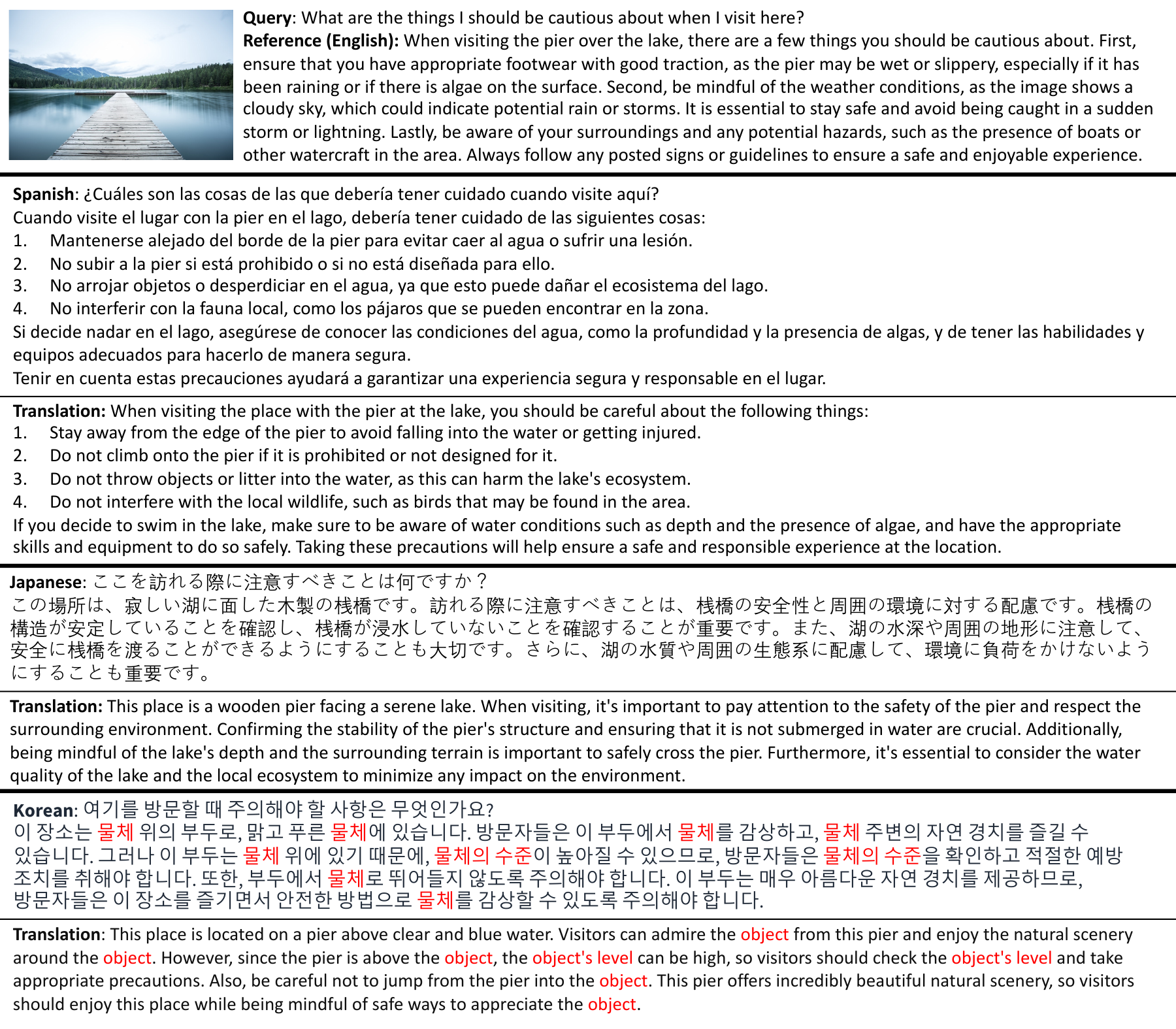} \\

\caption{\textbf{Compositional capability: multilingual visual conversation.} \newshortname{} generalizes to  multilingual visual conversations, when training on visual instruction following data (English-only) together with the text-only ShareGPT data (multilingual). However, there can still be errors in some languages (\eg Korean, errors marked in {\color{red} red}).}
\label{fig:multilingual_visual}
\end{figure*}

\clearpage

\subsection{Compositional Capabilities}
\label{sec:appendix_compositional_capabilities}

We present qualitative examples of the compositional capabilities of \newshortname{}.
As shown in Fig.~\ref{fig:multilingual_visual}, \newshortname{} is capable of participating in multilingual visual conversations and adapting its output language based on the user's input, even though it has not been trained on multilingual visual instruction data. We hypothesize this emerging bahavior is a compositional capability learned from visual conversations (English-only) and the text-only ShareGPT data (multilingual). However, there can still be errors in some languages (\eg Korean), which could be improved by incorporating more of those language data.

Additionally, in Table~\ref{tab:visual_writing_task}, we show another observed compositional capability after including the ShareGPT data in training. \newshortname{} is able to produce more detailed and visually-grounded responses in writing tasks with visual inputs than \shortname{}.

\begin{table}[h!]
  \begin{minipage}{0.99\linewidth}
\centering
\scalebox{0.88}{
\begin{tabular}{p{1.7cm} p{6.7cm} }
\toprule
 \multicolumn{2}{l}{\bf Visual input example, Constrained SD Prompt Generation:}  \\
\midrule
&  \includegraphics[height=5cm]{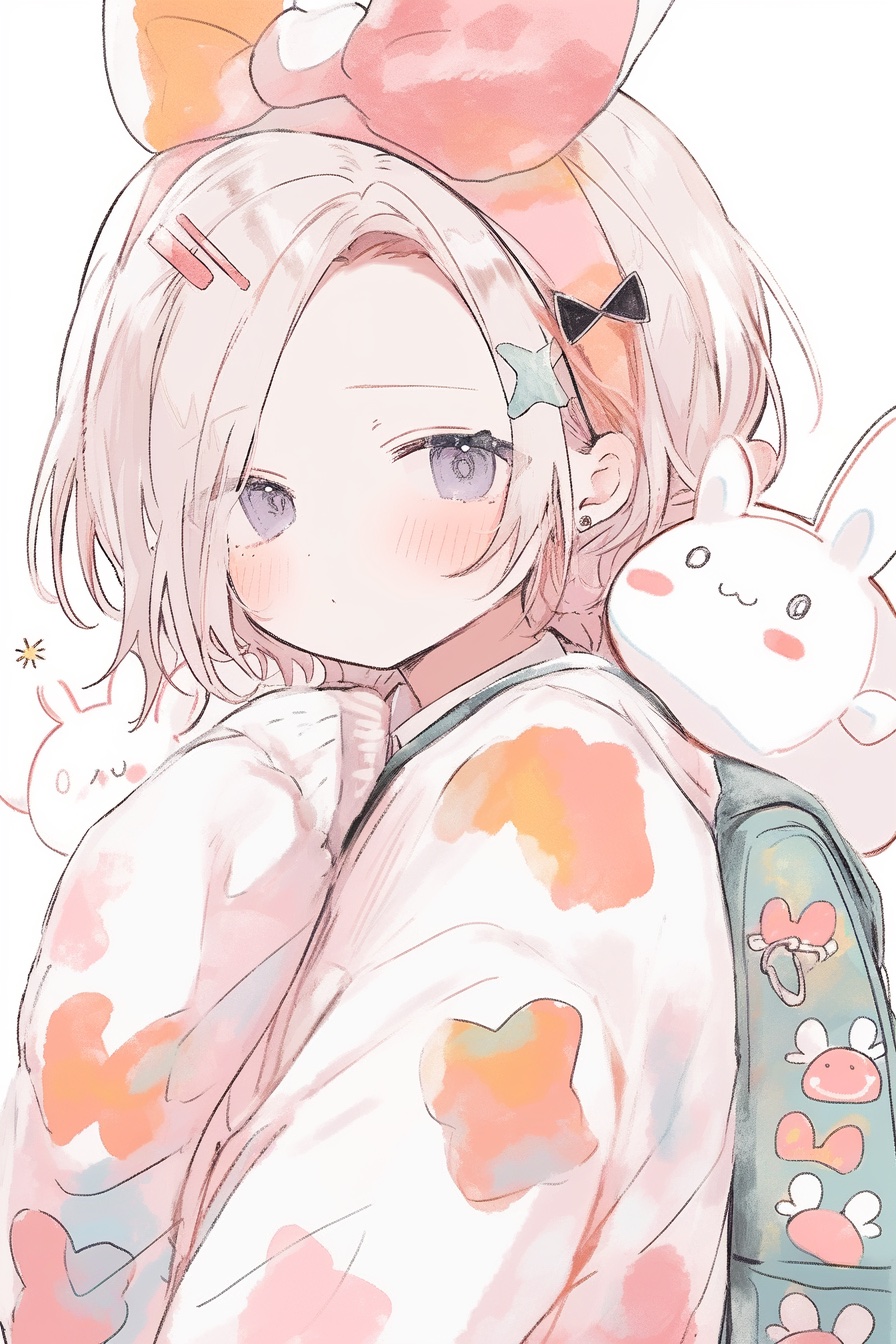} \\

User & Here's a formula for a Stable Diffusion image prompt: a cartoon-style image of {\color{orange}[adjective]} {\color{blue}[subject]} {\color{purple}[haircut]} {\color{brown}[clothing]} {\color{red}[doing action]}, {\color{emerald}[color scheme]}, detailed, trending on artstation. Generate the prompt for this image. \\
\newshortname{} & A cartoon-style image of {\color{orange}a young} {\color{blue}girl} {\color{purple}with a bob haircut}, {\color{brown}wearing a white kimono}, {\color{red}holding a stuffed animal, and posing for a picture}, {\color{emerald}with a color scheme of pink and white}, detailed, and trending on artstation. \\
\bottomrule
\end{tabular}
}
\captionof{table}{Constrained prompt generation for Stable Diffusion. Corresponding components are marked in {\color{orange}c}{\color{blue}o}{\color{purple}l}{\color{brown}o}{\color{red}r}.}
\label{tab:anime_prompt_generation}
  \end{minipage}
\end{table}

\section{Limitations}
\label{sec:limitation}

Despite the promising results demonstrated by \newshortname{}, several limitations must be acknowledged.
First, \newshortname{} utilizes full image patches, potentially prolonging each training iteration. While visual resamplers~\cite{li2023blip,dai2023instructblip,bai2023qwen} reduce the number of visual patches in LLMs, they currently cannot achieve convergence as efficiently as LLaVA with a comparable amount of training data, probably due to more trainable parameters in the resamplers. The development of a sample-efficient visual resampler could pave the way for future scaling-up of instruction-following multimodal models. 
Second, \newshortname{} is not yet capable of processing multiple images due to the lack of such instruction-following data, and the limit of the context length.
Third, although \newshortname{} exhibits proficiency in following complex instructions, its problem-solving capabilities can still be limited in certain domains, which could be improved with a more capable language model and with high-quality, targeted visual instruction tuning data.
Finally, despite its significantly reduced propensity for hallucination, \newshortname{} is not exempt from producing hallucinations and occasionally disseminating misinformation, and should be used with caution in critical applications (\eg medical).



{
    \small
    \bibliographystyle{ieeenat_fullname}
    \bibliography{main}
}

\end{document}